\documentclass[conference,a4paper]{IEEEtran}
\IEEEoverridecommandlockouts

\usepackage[hidelinks]{hyperref}
\usepackage[cmex10]{amsmath}
\usepackage{amssymb,amsfonts}
\usepackage{dblfloatfix}

\usepackage[ruled,vlined]{algorithm2e}
\usepackage{graphicx}
\graphicspath{{Figures/PDF/}{Figures/PNG/}}

\usepackage{booktabs}
\usepackage{siunitx}
\usepackage[numbers,compress]{natbib}
\usepackage{texnames}
\usepackage{bm,bbm}
\usepackage{orcidlink}
\usepackage{multirow}
\usepackage{xcolor}
\definecolor{bestColor}{RGB}{255, 0, 0}    
\definecolor{secondBestColor}{RGB}{0, 0, 255} 
\definecolor{thirdBestColor}{RGB}{240,0, 240}  
\definecolor{qBestColor}{RGB}{0,215, 0}  

\definecolor{background_color}{RGB}{255,255,255}  

\definecolor{cnn_color}{RGB}{240,248,235}  
\definecolor{transformer_color}{RGB}{245,240,230}  
\definecolor{mamba_color}{RGB}{235,235,245}  
\definecolor{ours_color}{RGB}{255,225,240}  
\definecolor{input_color}{RGB}{255,240,240}      

\usepackage{colortbl}  
\definecolor{bestColor}{RGB}{255, 0, 0}    
\definecolor{secondBestColor}{RGB}{0, 0, 255} 
\definecolor{thirdBestColor}{RGB}{0, 100, 0}  

\newcommand{\best}[1]{\textcolor{bestColor}{\textbf{#1}}}      
\newcommand{\secondBest}[1]{\textcolor{secondBestColor}{\textbf{#1}}} 
\definecolor{darkgreen}{rgb}{0.0, 0.5, 0.0} 

\begin{document}
\title{Robust Hyperspectral Image Panshapring via \\Sparse Spatial-Spectral Representation
\thanks{\textit{(Corresponding author: Chih-Chung Hsu; Email: cchsu@gs.ncku.edu.tw)}}
}

\author{\IEEEauthorblockN{Chia-Ming Lee\orcidlink{0009-0004-6027-3083}\IEEEauthorrefmark{1}, Yu-Fan Lin\orcidlink{0009-0000-9459-701X}\IEEEauthorrefmark{1}, Li-Wei Kang\orcidlink{0000-0001-6529-3981}\IEEEauthorrefmark{2}, Chih-Chung Hsu\orcidlink{0000-0002-2083-4438}\IEEEauthorrefmark{1}}\IEEEauthorblockA{\IEEEauthorrefmark{1}\textit{National Cheng Kung University}, Tainan, Taiwan (R.O.C.)\\\{zuw408421476@gmail.com, aas12as12as12tw@gmail.com, cchsu@gs.ncku.edu.tw\}}\IEEEauthorblockA{\IEEEauthorrefmark{2}\textit{National Taiwan Normal University}, Taipei, Taiwan (R.O.C.)\\\{lwkang@ntnu.edu.tw\}}}

\maketitle

\begin{abstract}
High-resolution hyperspectral imaging plays a crucial role in various remote sensing applications, yet its acquisition often faces fundamental limitations due to hardware constraints. This paper introduces S$^{3}$RNet, a novel framework for hyperspectral image pansharpening that effectively combines low-resolution hyperspectral images (LRHSI) with high-resolution multispectral images (HRMSI) through sparse spatial-spectral representation.
The core of S$^{3}$RNet is the Multi-Branch Fusion Network (MBFN), which employs parallel branches to capture complementary features at different spatial and spectral scales. Unlike traditional approaches that treat all features equally, our Spatial-Spectral Attention Weight Block (SSAWB) dynamically adjusts feature weights to maintain sparse representation while suppressing noise and redundancy. To enhance feature propagation, we incorporate the Dense Feature Aggregation Block (DFAB), which efficiently aggregates inputted features through dense connectivity patterns. This integrated design enables S$^{3}$RNet to selectively emphasize the most informative features from differnt scale while maintaining computational efficiency.
Comprehensive experiments demonstrate that S$^{3}$RNet achieves state-of-the-art performance across multiple evaluation metrics, showing particular strength in maintaining high reconstruction quality even under challenging noise conditions. The code will be made publicly available.
\end{abstract}


\section{Introduction}

Hyperspectral imaging (HSI) has revolutionized remote sensing by enabling precise material identification and classification through its ability to capture detailed spectral information across hundreds of contiguous bands \cite{ghamisi2017advances, Vivone2023multispectral}. However, acquiring high-resolution hyperspectral images (HRHSIs) faces fundamental challenges due to hardware limitations and spatial-spectral trade-offs \cite{mei2022hyperspectral}. While image fusion techniques, particularly pansharpening, offer a promising solution by combining low-resolution hyperspectral images (LRHSI) with high-resolution multispectral images (HRMSI) or Panchromatic (PAN) images, traditional approaches including wavelet-based \cite{Nunez1999multiresolution} and sparse representation methods \cite{dong2016hyperspectral} often struggle with real-world challenges such as noise and varying illumination conditions. These limitations further impede HSI-related real-world applications \cite{rtcs, lee2024prompthsiuniversalhyperspectralimage}.

\begin{figure}
    \centering
    \includegraphics[width=0.5\textwidth]{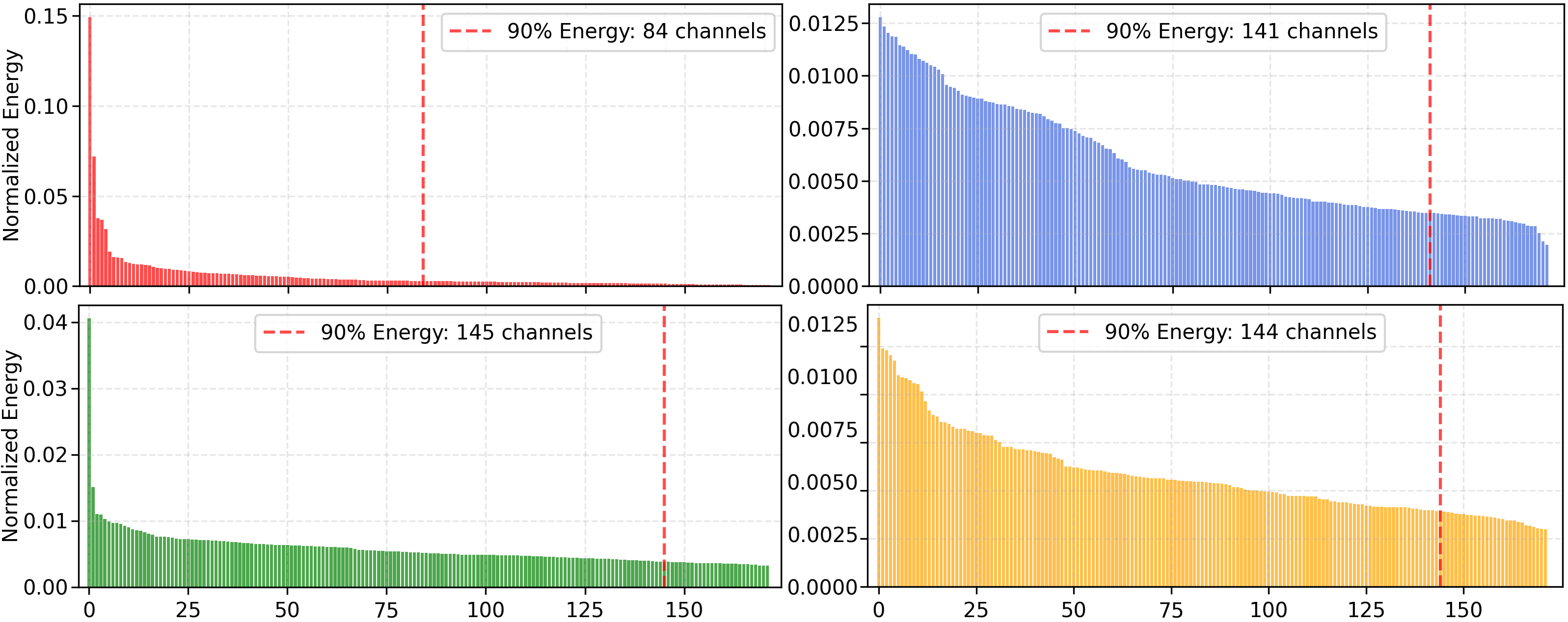}
    \vspace{-0.75cm}   
    \caption{Normalized energy distribution of fused layer across 172 channels between \textcolor{red}{the proposed S$^{3}$RNet}, \textcolor{blue}{CNN-based method} \cite{Min2021MSSJFL}, \textcolor{darkgreen}{Transformer-based method} \cite{HyperTransformer}, and \textcolor{orange}{Mamba-based method} \cite{FusionMamba}. The majority of energy concentrate in head side, indicating our method's sparsity. By integrating spatial-spectral representation, our method have archived state-of-the-art performance with robustness against heavy noise compared with other peer-methods.}\vspace{-0.2cm} 
    \label{fig:de}\vspace{-0.2cm} 
\end{figure}

Recent advances in deep learning have introduced data-driven approaches that significantly improve fusion performance. Architectures such as convolutional neural networks (CNNs) \cite{Zhu2021PZRes,Min2021MSSJFL,U2Net,DHIF-Net,Xiao2021DualUNet}, Transformers \cite{fusformer,HyperRefiner,HyperTransformer}, and Mamba-based methods \cite{FusionMamba} have shown promising results in modeling spatial-spectral correlation and integration. However, these methods often rely on dense feature representations that can lead to redundancy and increased computational complexity.

The concept of sparsity plays a crucial role in hyperspectral image processing, particularly in enhancing noise robustness and computational efficiency. Recent studies \cite{dong2016hyperspectral,sparse1,sparse2} have demonstrated that sparse representations can effectively capture essential spatial-spectral patterns while inherently suppressing noise interference for HSI reconstruction tasks. 

On the other hand, existing methods often employ fixed feature integration strategies that fail to adapt to varying input and feature characteristics. As demonstrated in Figure \ref{fig:de}, existing CNN, Transformer, and Mamba-based methods tend to produce smooth and uniform energy distributions across spectral channels, indicating their inability to identify and prioritize the most informative features. In contrast, our approach achieves significantly higher channel-wise sparsity with energy concentrated in key spectral bands, enabling more effective noise suppression while preserving essential spectral characteristics. 

In this paper, we propose S$^{3}$RNet, a novel framework that seamlessly integrates sparse representation with multi-scale learning for hyperspectral image pansharpening. By combining complementary spatial-spectral features at multiple scales with adaptive feature weighting that dynamically adjusts feature sparsity across different spatial-spectral scales based on multi-branches architecture. Afterwards, sparse representation aligns with the inherent structure of hyperspectral data, where meaningful information is often concentrated in specific spectral regions rather than uniformly distributed across all channels. Thus, the high-quality HSI can be fused even noisy source inputs. In summary, our approach achieves superior pansharping performance while maintaining robustness against noise. Our contribution can be listed as threefold:

\begin{itemize}
\item \textbf{Multi-Branch Architecture}: The proposed Multi-Branch Fusion Network (MBFN) captures complementary features through parallel branches operating at different spatial-spectral scales. These branches employ upsampling and downsampling operations with convolutional layers, naturally suppressing high-frequency noise while preserving essential spatial-spectral details.

\item \textbf{Weighted Sparse Representation}: Our Spatial-Spectral Attention Weight Block (SSAWB) adaptively adjusts feature importance across branches based on input characteristics. This dynamic mechanism promotes sparse representation by emphasizing informative features while suppressing redundant or noisy ones, enabling robust fusion across diverse scenarios.

\item \textbf{Dense Feature Integration}: We design an efficient Dense Feature Aggregation Block (DFAB) that enhances feature reuse through dense connectivity. This module not only improves computational efficiency but also contributes to noise resilience through implicit feature refinement, ensuring stable training and high-quality reconstruction.
\end{itemize}

\section{PROPOSED METHOD}\label{sec:theory}
\subsection{Architecture Overview}

Hyperspectral image pansharpening aims to fuse LRHSI with HRMSI or PAN image to achieve both high spatial and spectral resolution. However, traditional fusion methods often struggle with noise sensitivity and computational inefficiency, leading to degraded reconstruction quality.

\begin{figure*}[htbp]
\centering
\includegraphics[width=1\linewidth]{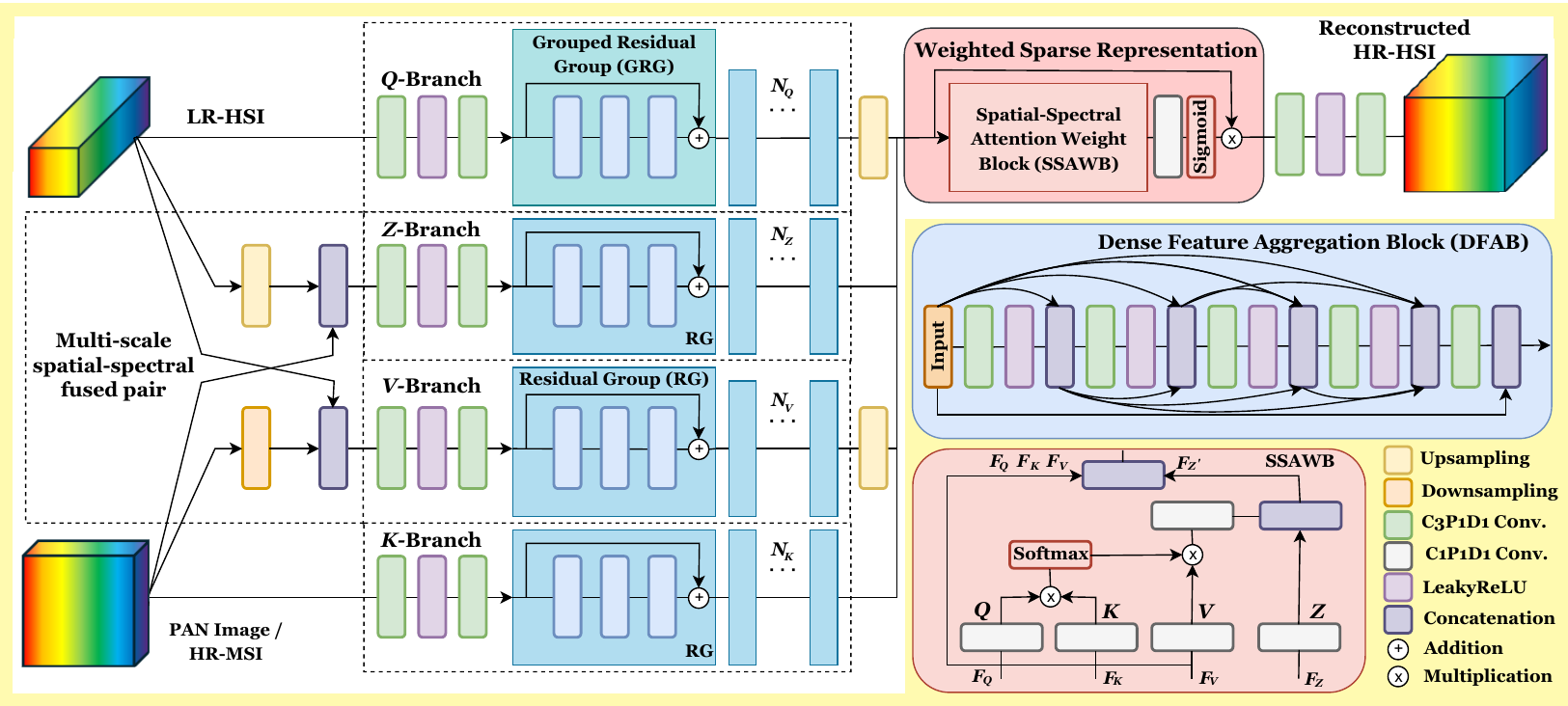}\vspace{-0.25cm}
\caption{{\textbf{Architecture of the proposed S$^3$RNet for hyperspectral image pansharping.} Multi-Branch Fusion Network (MBFN) for extracting multi-scale spatial-spectral features in $\mathbf{Q}$-$\mathbf{K}$-$\mathbf{V}$-$\mathbf{Z}$ branch, Spatial-Spectral Attention Weight Block (SSAWB) for adaptive feature refinement and improving robustness against noise interference with sparse feature representation, and Dense Feature Aggregation Block (DFAB) for efficient feature integration.}}\vspace*{-0.6cm}
\label{fig:net}
\end{figure*}

In this manuscript, we proposed S$^3$RNet to addresses these challenges, (the overall architecture have illustrated in Figure \ref{fig:net}) through two major concepts: \textbf{sparse representation} and \textbf{multi-scale spatial-spectral learning}. The proposed MBFN captures complementary features across different spatial-spectral scales for feature extraction. The SSAWB further enhances noise resilience by dynamically weighting these extracted features to maintain a sparse representation according to dual-inputs adaptively, enabling efficient fusion while preserving both spatial and spectral fidelity.

This architecture enables S$^3$RNet to achieve superior pansharpening performance by effectively integrating spatial details from HRMSI with rich spectral information from LRHSI, while maintaining robustness against noise and computational efficiency through its emphasis on sparse representation.

\begin{figure}
    \centering
    \includegraphics[width=0.49\textwidth]{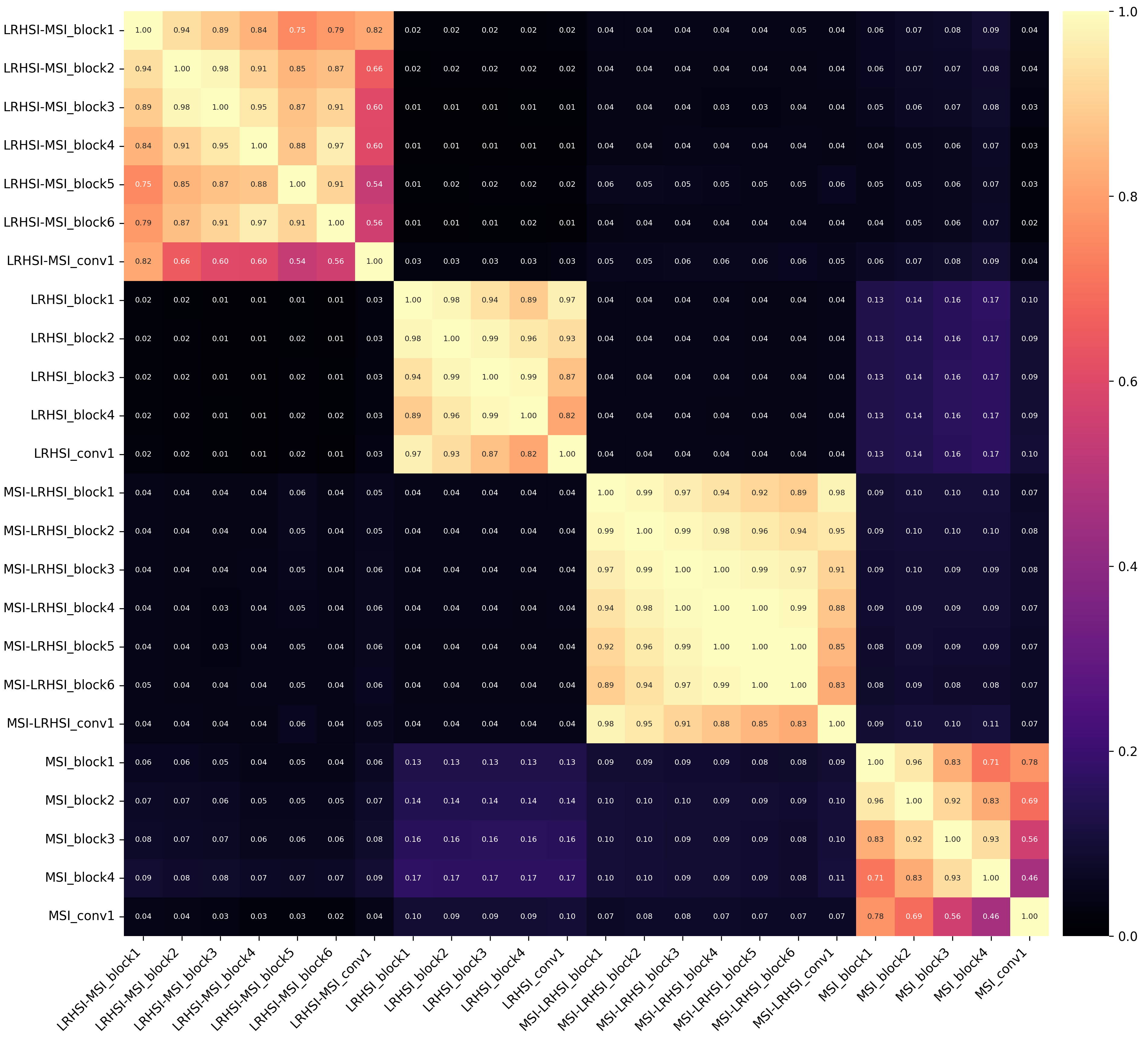}
    \vspace{-0.75cm}   
    \caption{Layer-wise CKA similarity analysis \cite{CKA}. It indicates that different branches ($\mathbf{Q}$-$\mathbf{K}$-$\mathbf{V}$-$\mathbf{Z}$ branch) learn specialized feature representations. The low cross-branch similarity (dark regions) suggests each branch captures unique aspects of the spatial-spectral information, rather than redundancy.}\vspace{-0.6cm} 
    \label{fig:cka} 
\end{figure}

\subsection{Multi-Scale Spatial-Spectral Representation Learning}

\textbf{Multi-scale Spatial-Spectral Feature Extraction.} Compared with traditional single-branch or dual-branch designs, the multi-branch architecture of MBFN offers several advantages. \textbf{First}, the multi-branch architecture promotes information sharing and specialization, enabling branches to focus on distinct aspects of the input data, as shown in the Figure \ref{fig:cka}. Therefore, it could ensures effective learning of spatial-spectral features across multiple granularities, enhancing the quality of reconstructed hyperspectral images. Formally, our MBFN can be defined as following process:

Let the LRHSI, HRMSI (or PAN), HRHSI (ground-truth) be denoted as $\mathbf{X}_{h}$, $\mathbf{X}_m$, and $\mathbf{Y}$, respectively. The LRHSI can be expressed as $\mathbf{X}_h = \mathbf{Y}\mathbf{B}$, where $\mathbf{B}$ represents a blurring matrix that reduces spatial resolution by aggregating neighboring pixels. Similarly, the HRMSI is modeled as $\mathbf{X}_m = \mathbf{D}\mathbf{Y}$, where $\mathbf{D}$ is a downsampling matrix that reduces the spectral resolution by combining adjacent bands, $\mathcal{G}$($\cdot$) denotes feature transition function \cite{memnet}, $\textbf{Cat.}$($\cdot$) represents channel-wise concatenation. 

The MBFN contains four branches, $\mathbf{Q}$-$\mathbf{K}$-$\mathbf{V}$-$\mathbf{Z}$ branch, respectively, for multi-scale spatial-spectral representation learning, as shown in Figure \ref{fig:net}, can be simply defined as:\vspace{-0.1cm}
\begin{equation}
\begin{aligned}
\mathbf{F}_{Q} &= \text{GRG}_{i}^{Q}(...(\text{GRG}_{1}^{Q}(\mathcal{G}^{Q}(\mathbf{X}_h)))),\quad i = 1, 2, \ldots, N_{Q}; \\
\mathbf{F}_{K} &= \text{RG}_{i}^{K}(...(\text{RG}_{1}^{K}(\mathcal{G}^{K}(\mathbf{X}_m)))),\quad i = 1, 2, \ldots, N_{K}, \\
\mathbf{F}_{V} &= \text{RG}_{i}^{V}(...(\text{RG}_{1}^{V}(\mathcal{G}^{V}(\textbf{Cat.}(\mathbf{X}_m^d, \mathbf{X}_h))))),\quad i = 1, 2, \ldots, N_{V}; \\
\mathbf{F}_{Z} &= \text{RG}_{i}^{Z}(...(\text{RG}_{1}^{Z}(\mathcal{G}^{Z} (\textbf{Cat.}(\mathbf{X}_h^u, \mathbf{X}_m))))),\quad i = 1, 2, \ldots, N_{Z}; \\
\end{aligned}\vspace{-0.08cm}
\end{equation}
where GRG($\cdot$) and RG($\cdot$) represent a Grouped Residual Group (GRG) and Residual Group (RG), a core block of each branches for feature extraction, consisting of consecutive DFABs, we will introduce these DFAB in details later. By leveraging the low-rank prior of HSI \cite{lowrankprior1,lowrankprior2}, GRG replaces standard convolution with group convolution to further reduce parameter-size with negligible performance drops. Note that we set $N_{Q}$, $N_{K}$, $N_{V}$, $N_{Z}$ to 4, 4, 6, 6, respectively.

\textbf{Second}, maintaining the network's efficiency is also essential in MBFN architecture. Using wide and shallow multi-branch networks also helps to improve the parallelization and low-latency of feature extraction. By combining these design principles, the MBFN framework captures a comprehensive spatial-spectral feature representation in S$^3$RNet while maintaining computational efficiency for HSI pansharping.

\subsection{Dense Feature Aggregation Block}
\textbf{Effective Information Propagation.} The proposed DFAB serves as the foundational building unit of RG/GRG in the MBFN framework, designed to enhance spatial-spectral representation for information propagation. Due to the information bottleneck when the depth of residual-in-residual network increases \cite{Hsu_2024_CVPR}, the representation capability of reconstruction tasks, such as fusion \cite{lee2025hyfusionenhancedreceptionfield} or super-resolution \cite{Chen_2024_CVPR,Ren_2024_CVPR,Zhang_2023_CVPR}, would be limited. Through dense connections, DFAB facilitates efficient information flow, and stabilizes the learning process.

The dense connection in DFAB progressively accumulates and reuses features with the input feature $\mathbf{F}_{\text{in}}$ for feature aggregation, $\sigma$ denotes LeakyRELU activation with slope $0.2$:
\vspace{-0.1cm} 
\begin{equation}
\begin{aligned}
\mathbf{F}_j &= \textbf{Cat.}(\sigma^{j}(\text{Conv}_{1\times1}^{j}(\mathbf{F}_{\text{in}}, ..., \mathbf{F}_{j-1}))), j = 1, 2, 3, 4  \\
\mathbf{F}_{\text{out}} &= 0.2 \times \text{Conv}_{1\times1}^{5}(\mathbf{F}_4)+\mathbf{F}_{\text{in}}
\end{aligned}
\vspace{-0.1cm} 
\end{equation}

By combining dense connectivity patterns with efficient feature propagation, DFAB serves as a critical component that strengthens the entire MBFN architecture. DFAB enables comprehensive feature reuse while maintaining computational efficiency, allowing it to effectively suppress noise artifacts through layer-by-layer implicit feature refinement.

\begin{table*}[ht]
\centering
\caption{{\textbf{Performance evaluation and complexity comparison of the proposed S$^{3}$RNet and other HSI pansharping methods.}} The best results are highlighted in \best{bold-red}, while the second-best results are highlighted in \secondBest{bold-blue}. Rows are colored to distinguish different approaches: \colorbox{cnn_color}{CNN-based}, \colorbox{transformer_color}{Transformer-based}, \colorbox{mamba_color}{Mamba-based}, and \colorbox{ours_color}{the proposed} methods.}
\label{tab:performance}\vspace{-0.25cm} 
\scalebox{0.83}{
\begin{tabular}{r|cccc|cccc|cccc}  
\toprule[0.15em]
\multirow{2}{*}{\textbf{Method}} &  \multicolumn{4}{c|}{Model Complexity} & \multicolumn{4}{c|}{4 Bands LR-HSI} & \multicolumn{4}{c}{6 Bands LR-HSI}\\
 &  \#Params↓ & FLOPs↓ & Run-time↓ & Memory↓ & PSNR↑ &  SAM↓ &  RMSE↓  & ERGAS↓  &  PSNR↑ &  SAM↓ &  RMSE↓ & ERGAS↓   \\\hline
\rowcolor{cnn_color}
\textbf{PZRes-Net} \cite{Zhu2021PZRes} {\tt\small{TIP'21}} & 40.15M & 5262.34G & 0.0141s & 11059MB & 34.963 &  1.934  & 35.498 & 1.935& 37.427 &  1.478 &  28.234 &  1.538  \\
\rowcolor{cnn_color}
\textbf{MSSJFL} \cite{Min2021MSSJFL} {\tt\small{HPCC'21}} & 16.33M & 175.56G & \secondBest{0.0128s} & \best{1349MB} &  34.966 & 1.792 &  33.636 &  2.245 &  38.006 &  1.390 &  26.893 & 1.535 \\
\rowcolor{cnn_color}
\textbf{Dual-UNet} \cite{Xiao2021DualUNet} {\tt\small{TGRS'21}} & \secondBest{2.97M} & \secondBest{88.65G} & \best{0.0127s} & \secondBest{2152MB} & \secondBest{35.423} & 1.892 & 33.183 & \secondBest{1.796} & 38.453 & 1.548 & 26.148 & 1.205 \\
\rowcolor{cnn_color}
\textbf{DHIF-Net} \cite{Huang2022DHIF} {\tt\small{TCI'22}} & 57.04M & 13795.11G &  6.005s & 29381MB &34.458 & 1.829 & 34.769 & 2.613 & \secondBest{39.146} & 1.239 & 25.309 & \secondBest{1.113} \\
\rowcolor{transformer_color}
\textbf{FusFormer} \cite{fusformer} {\tt\small{TGRS'22}} & \best{0.18M} & \best{11.74G} & 0.0158s & 5964MB & 34.217 & 2.012 & 35.687 & 1.996 & 38.637 & 1.678 & 28.674 & 1.204 \\
\rowcolor{transformer_color}
\textbf{HyperTransformer} \cite{HyperTransformer} {\tt\small{CVPR'22}} & 142.83M & 343.96G & 0.0252s & 8104MB & 28.692 &  3.664 &  62.231 & 4.774 & 32.954 & 2.568 & 41.256 & 3.834 \\
\rowcolor{transformer_color}
\textbf{HyperRefiner} \cite{HyperRefiner} {\tt\small{TJDE'23}} & 19.32M & 94.37G & 0.0237s & 7542MB & 33.298 & 2.129 & 38.769 & 2.086 & 37.654 & 1.590 & 29.629 & 1.403  \\
\rowcolor{cnn_color}
\textbf{U2Net} \cite{U2Net} {\tt\small{ACMMM'23}} & 265.15M & 1931.09G & 0.1684s & 7506MB & 25.622 & 3.855 & 86.682 & 7.341 & 27.068 & 3.832 & 85.101 & 6.816 \\
\rowcolor{cnn_color}
\textbf{QRCODE} \cite{QRCODE} {\tt\small{TGRS'24}} & 41.88M & 2231.19G & 0.2452s & 15028MB & 35.361 & \secondBest{1.623} & \secondBest{32.711} & 2.027 & 38.948 & \secondBest{1.148} &  \secondBest{24.617} & 1.429 \\ 
\rowcolor{mamba_color}
\textbf{FusionMamba} \cite{FusionMamba} {\tt\small{TGRS'24}} & 21.68M & 134.47G & 0.0347s & 2446MB & 30.741 & 1.978 &  50.744 & 2.945 & 32.407 & 1.540 & 45.774 & 2.569 \\\hline
\rowcolor{ours_color}
\textbf{S}$^{3}$\textbf{RNet (Ours)} & 26.81M & 941.77G &  0.0134s & 8733MB & \best{35.967} &  \best{1.527} &  \best{30.928} &  \best{1.554}& \best{40.046} & \best{1.095} &  \best{23.785} &  \best{1.092}  \\
\bottomrule[0.15em]
\end{tabular}}\vspace{-0.55cm} 
\end{table*}

\begin{table}[!t]

\caption{{\textbf{Robustness evaluation.}} 6 bands LR-HSI with the AWGN noise.}
\label{table:noise}
\vspace{-2.3mm}
\scalebox{0.9}{
\begin{tabular}{r|c|c}
\toprule[0.15em]
\multirow{2}{*}{\textbf{Method}} & SNR 35$\%$ (Cleaner) & SNR 15$\%$ (Noiser) \\ 
 & PSNR / SAM / ERGAS & PSNR / SAM / ERGAS\\ 
\hline
\rowcolor{cnn_color}
\textbf{PZResNet} \cite{Zhu2021PZRes}& 28.772 / 4.030 / 9.819 & 9.372 / 30.635 / 99.383 \\
\rowcolor{cnn_color}
\textbf{MSSJFL} \cite{Min2021MSSJFL}& 31.768 / 2.918 / 4.638 & 14.355 / 20.776 / 37.812 \\
\rowcolor{cnn_color}
\textbf{Dual-UNet} \cite{Xiao2021DualUNet}& 26.935 / 5.042 / 10.176& 7.910 / 35.722 / 101.951  \\
\rowcolor{cnn_color}
\textbf{DHIF-Net} \cite{Huang2022DHIF}& 27.424 / 4.512 / 10.124 & 9.104 / 28.264 / 93.581 \\
\rowcolor{transformer_color}
\textbf{FusFormer} \cite{fusformer}& 26.409 / 5.302 / 10.481 & 9.801 / 35.347 / 78.912 \\
\rowcolor{transformer_color}
\textbf{HyperTransformer} \cite{HyperTransformer}& 31.303 / 3.048 / 5.214 & \secondBest{23.525} / \secondBest{7.759} / \secondBest{10.519} \\
\rowcolor{transformer_color}
\textbf{HyperRefiner} \cite{HyperRefiner}& 23.095 / 7.064 / 27.066 & 11.226 / 25.763 / 77.412 \\
\rowcolor{cnn_color}
\textbf{U2Net} \cite{U2Net}& 24.459 / 4.322 / 8.715 & 15.043 / 14.484 / 38.927 \\
\rowcolor{cnn_color}
\textbf{QRCODE} \cite{QRCODE}& \secondBest{32.265} / 9.108 / \secondBest{2.642} & 10.516 / 27.966 / 69.524   \\
\rowcolor{mamba_color}
\textbf{FusionMamba} \cite{FusionMamba}& 28.921 / \secondBest{2.502} / 4.687 & 15.822 / 13.602 / 34.737 \\\hline
\rowcolor{ours_color}
\textbf{S}$^{3}$\textbf{RNet (Ours)}  & \best{36.411} / \best{1.785} / \best{2.351} & \best{27.682} / \best{4.767} / \best{6.046} \\
\bottomrule[0.15em]
\end{tabular}
}
\vspace{-0.5cm}
\end{table}

\subsection{Weighted Sparse Representation}
\textbf{Spatial-Spectral Attention for Feature Aggregation.} Tradition methods directly concate or add different modalities feature, which treat all branches equally, impeding the information interaction. To overcome this, the SSAWB employs a cross attention mechanism for adaptive multi-scale spatial-spectral information aggregation.

Given the features from MBFN with $\mathbf{Q}$-$\mathbf{K}$-$\mathbf{V}$-$\mathbf{Z}$ branches, $\mathbf{F}_{Q}$, $\mathbf{F}_{K}$,  $\mathbf{F}_{V}$, and $\mathbf{F}_{Z}$ with the linear projection layers into lower dimensional space $d$ with reduction parameter $r$, we can derive the projected feature $\mathbf{Q}$, $\mathbf{K}$, $\mathbf{V}$, $\mathbf{Z}$ for SSAWB. Now, we could perform the cross-attention: 
\vspace{-0.2mm}
\begin{equation}
\mathbf{Z}^{'} = \text{Conv}_{1\times1}(\textbf{Softmax}\left(\frac{\mathbf{Q} \cdot \mathbf{K}^T}{\sqrt{d_{r}}}\right)\mathbf{V})+\mathbf{Z}, \vspace{-0.2mm}
\end{equation}
\begin{equation}
\mathbf{C} = \text{Conv}_{1\times1}^{\text{adapt.}}(\text{Cat.}(\mathbf{Q}, \mathbf{K}, \mathbf{V}, \mathbf{Z}^{'})), \vspace{-0.2mm}
\end{equation}
where $\text{Conv}_{1\times1}^{\text{adapt.}}(\cdot)$ projects the concatenated cross-attentions to adaptive weights $\mathbf{W}$ for dynamic feature aggregation and reconstruction. In this way, we could judiciously fuse the spatial-spectral representation at various scale.

\textbf{Sparsity from Weighted Sigmoid Operator.} Afterward, we apply the Sigmoid function to determine the importance of each feature, enabling the model to assign independent attention scores, as follows:\vspace{-0.2mm}
\begin{equation}
\mathbf{W} = \textbf{Sigmoid}(\mathbf{C})
\end{equation}
\vspace{-1.4mm}
\begin{equation}
    \mathbf{F}_{\text{fused}} = \mathbf{W}_1 \cdot \mathbf{Q} + \mathbf{W}_2 \cdot \mathbf{K} + \mathbf{W}_3 \cdot \mathbf{V} + \mathbf{W}_4 \cdot \mathbf{Z}^{'},
\end{equation}
\vspace{-0.2mm}
Finally, the reconstructed HR-HSI is obtained via a simple feature transition function $\mathcal{G}$($\cdot$) by
$\mathbf{Y}^*$ = $\mathcal{G}$$(\mathbf{F}_{\text{fused}})$.

The Sigmoid activation enables independent evaluation of each branch's importance, promoting feature sparsity while effectively suppressing noise. Unlike Softmax which can either over-emphasize or completely ignore features, Sigmoid's non-competitive nature makes it ideal for multi-branch fusion by preserving unique contributions from each branch. Through this weighted attention mechanism, SSAWB achieves robust feature integration that maintains high reconstruction quality even under challenging conditions.

\section{Experimental Results}\label{sec:expALL}

\subsection{Experiment Settings}

\subsubsection{Dataset Illustration}
The dataset used for performance evaluation in this study was acquired by the Airborne Visible/Infrared Imaging Spectrometer (AVIRIS) sensor \cite{Vane1993airborne}. It consisted of 2,078 HR-HSI images, which were randomly partitioned into training, validation, and testing sets for performance evaluation. The training set contained 1,678 images, while the validation and testing sets contained 200 images for each. The spatial and spectral resolutions of the HR-MSI and LR-HSI were $256\times 256 \times M_m$ and $64\times 64\times 172$, respectively, where $M_m$ is either 4 or 6 in our experiments.

\subsubsection{Implementation Details} All experiments are conducted on single NVIDIA-A100. The batch size was set to 4, and the number of training epochs was fixed to 600 for all experiments involving the proposed method. For the peer methods, the number of hyperparameters was set according to their default settings as specified in their original publications. The ADAM \cite{Kingma2014adam} optimizer was used for training, with an initial learning rate of 0.0001. The learning rate was adjusted with cosine annealing scheduler. Standard augmentation, including random cropping, flipping and rotation, are employed.

\subsubsection{Task-Specific Optimization}
To improve spectral fidelity in hyperspectral image reconstruction, we incorporate the Spectral Angle Mapper (SAM) as an auxiliary loss function alongside the standard L1 loss. The total loss is defined as a weighted combination of L1 and SAM losses, with SAM weighted at 0.1. This combined approach helps preserve both spatial and spectral characteristics of the reconstructed images.

\subsection{Performance Evaluation}

We evaluated our approach against seven state-of-the-art hyperspectral image fusion methods: PZRes-Net \cite{Zhu2021PZRes}, MSSJFL \cite{Min2021MSSJFL}, Dual-UNet \cite{Xiao2021DualUNet} and DHIF-Net \cite{Huang2022DHIF}, FusFormer \cite{fusformer}, HyperTransformer \cite{HyperTransformer}, HyperRefiner \cite{HyperRefiner}, U2Net \cite{U2Net}, QRCODE \cite{QRCODE}, and FusionMamba \cite{FusionMamba}. The performance was objectively measured using PSNR, SAM, and RMSE, ERGAS. Experiments were conducted with both 4 and 6 MSI bands. As more bands in the LR-HSI are available, richer spectral information can be exploited to potentially enhance the SR quality. The quantitative results are presented in Table \ref{tab:performance}, demonstrating the proposed framework outperforms the compared state-of-the-art methods in terms of spectral reconstruction performance.

\subsection{Robustness Evaluation}
To assess model robustness, we tested performance under noise conditions by adding Additive White Gaussian Noise (AWGN) at two Signal-to-Noise Ratios (SNR): 35\% and 15\%. The AWGN noise was applied to 6 bands LRHSI for pansharping and evaluation. 

The experimental results, shown in Table \ref{table:noise}, demonstrate that while conventional methods exhibited performance degradation under severe noise conditions, S$^3$RNet maintained consistent performance, indicating superior robustness to signal interference.

\section{Conclusion}\label{sec:Conclusion}
In this work, we have introduced S$^3$RNet, a novel framework for hyperspectral image pansharping that effectively bridges the spatial and spectral resolution gap between LRHSI and HRMSI. Leveraging its Multi-Branch Fusion Network (MBFN), Spatial-Spectral Attention Weight Block (SSAWB), and Dense Feature Aggregation Block (DFAB), S$^3$RNet captures multi-scale spatial-spectral features with exceptional noise resilience and sparsity in representation. Experimental results demonstrate that S$^3$RNet outperforms state-of-the-art methods, delivering high-fidelity reconstruction even under challenging noise conditions, thus establishing a new benchmark for hyperspectral image enhancement.

\bibliographystyle{IEEEtran}
\bibliography{ref}

\end{document}